\documentclass{sigchi}

\CopyrightYear{2020}
\setcopyright{acmlicensed}
\doi{https://doi.org/10.1145/3313831.3376318}
\isbn{978-1-4503-6708-0/20/04}
\conferenceinfo{CHI'20,}{April  25--30, 2020, Honolulu, HI, USA}
\acmPrice{\$15.00}

\usepackage{balance}       
\usepackage{graphics}      
\usepackage[T1]{fontenc}   
\usepackage{txfonts}
\usepackage{mathptmx}
\usepackage[pdflang={en-US},pdftex]{hyperref}
\usepackage{color}
\usepackage{booktabs}
\usepackage{textcomp}
\usepackage{multirow}
\usepackage{enumitem}
\usepackage{microtype}        
\usepackage{ccicons}          

\usepackage{todonotes}

\def\plaintitle{Studying the Effects of Cognitive Biases in Evaluation of Conversational Agents}

\def\emptyauthor{}
\def\plainkeywords{Authors' choice; of terms; separated; by
  semicolons; include commas, within terms only; this section is required.}

\makeatletter
\def\url@leostyle{%
  \@ifundefined{selectfont}{
    \def\UrlFont{\sf}
  }{
    \def\UrlFont{\small\bf\ttfamily}
  }}
\makeatother
\def\pprw{8.5in}
\def\pprh{11in}

\definecolor{linkColor}{RGB}{6,125,233}
\hypersetup{%
  pdftitle={\plaintitle},
  pdfauthor={\emptyauthor},
  pdfkeywords={\plainkeywords},
  pdfdisplaydoctitle=true, 
  bookmarksnumbered,
  pdfstartview={FitH},
  colorlinks,
  citecolor=black,
  filecolor=black,
  linkcolor=black,
  urlcolor=linkColor,
  breaklinks=true,
  hypertexnames=false
}

\toappear{\scriptsize Permission to make digital or hard copies of all or part of this work for personal or classroom use is granted without fee provided that copies are not made or distributed for profit or commercial advantage and that copies bear this notice and the full citation on the first page. Copyrights for components of this work owned by others than ACM must be honored. Abstracting with credit is permitted. To copy otherwise, or republish, to post on servers or to redistribute to lists, requires prior specific permission and/or a fee. Request permissions from permissions@acm.org. \\
{\emph{CHI '20, April 25--30, 2020, Honolulu, HI, USA.} } \\
Copyright is held by the owner/author(s). Publication rights licensed to ACM. \\
ACM ISBN 978-1-4503-6708-0/20/04\ ...\$15.00.\\
http://dx.doi.org/10.1145/3313831.3376318}

\begin{document}

\title{\plaintitle}

\numberofauthors{1}
\author{%
  \alignauthor{Sashank Santhanam, Alireza Karduni and Samira Shaikh\\
    \affaddr{University of North Carolina at Charlotte}\\
    \affaddr{Charlotte, USA}\\
    \email{\{ssantha1,akarduni,samirashaikh\}@uncc.edu}}\\
}

\maketitle

\begin{abstract}
Humans quite frequently interact with conversational agents.  The rapid advancement in generative language modeling through neural networks has helped advance the creation of intelligent conversational agents. Researchers typically evaluate the output of their models through crowdsourced judgments, but there are no established best practices for conducting such studies. Moreover, it is unclear if cognitive biases in decision-making are affecting crowdsourced workers' judgments when they undertake these tasks. To investigate, we conducted a between-subjects study with 77 crowdsourced workers to understand the role of cognitive biases, specifically anchoring bias, when humans are asked to evaluate the output of conversational agents.  Our results provide insight into how best to evaluate conversational agents. We find increased consistency in ratings across two experimental conditions may be a result of anchoring bias. We also determine that external factors such as time and prior experience in similar tasks have effects on inter-rater consistency.
\end{abstract}




\section{Introduction}
Conversational agents, also commonly known as chatbots, are typically designed with the intention of generating meaningful, informative and coherent responses that keep humans engaged in conversation. Conversational agents have become extremely popular and have been heralded as one of the recent breakthrough technologies.\footnote{\url{https://www.technologyreview.com/lists/technologies/2016/}} The development of conversational agents has evolved from simple rule-based approaches such as Eliza \cite{weizenbaum1966eliza} and PARRY \cite{colby1975artificial} to more sophisticated templates-based \cite{mcroy2003augmented,van2005real} and data-driven approaches \cite{langkilde2002halogen,Bangalore:2000:CLC:1075218.1075277}. 
Extant approaches towards building conversational agents are end-to-end systems that employ \textit{seq2seq} architectures \cite{vinyals2015neural,sutskever2014sequence}, \textit{language modeling} \cite{bengio2003neural,mei17} or \textit{transformer} architectures \cite{vaswani2017attention}.

Even with the rapid advancement in the \textit{development} of conversational agents through these neural approaches, there are no established set of best practices towards \textit{evaluating} their performance. Evaluation procedures vary from one research article to the next, leading to a fragmented view of how the field is advancing. Overall, the output generated from these models is evaluated using automated metrics and/or (crowdsourced) human judgments. With respect to automated metrics, measures including BLEU \cite{papineni2002bleu}, METEOR \cite{banerjee2005meteor}, ROUGE \cite{lin2004rouge} and word-embedding based metrics \cite{dialogue-eval}, which can be calculated based on word overlap have been used. However, prior research has shown that these metrics show little to no correlation with human ratings \cite{novikova-etal-2017-need,lowe-etal-2017-towards,dialogue-eval}. Due to these limitations of automated metrics, evaluation of chatbots is increasingly conducted by obtaining qualitative judgments from crowd-sourced workers \cite{mei17,zhang-etal-2018-personalizing}. This puts a major imperative on how the experiments to collect crowd-sourced judgments are designed. However, research advancing best practices for experiment design for evaluating chatbots performance and obtaining more reliable and consistent ratings from crowd-sourced workers is very limited. Our work seeks to fill this research gap. 

Consider the simple choice of the type of question used to elicit human judgments. Most current experiments for evaluating conversational agent output use Likert scales; a typical question would be to ask the humans to rate the Readability of chatbot output on a scale of 1\textendash5. However, research by Belz and Kow \cite{belz2011discrete} has shown that using Likert scales may affect rating consistency, for example, some individuals may tend to avoid the extremes of the scale while others may not. Novikova \emph{et al.}  \cite{novikova-etal-2018-rankme} have shown that continuous scales help improve the consistency and reliability of human ratings across several language evaluation tasks as opposed to Likert scales. In their experiments, Novikova \emph{et al.} \cite{novikova-etal-2018-rankme} found that consistency of crowd-sourced workers improved when workers were asked to rate the conversational agent output by comparing it against a given (gold) standard. A sample question in their study would ask human raters to input a number to rate the Readability of an algorithm output by comparing it against the provided gold standard response (with a standard response value of 100). But what if this increased consistency is a result of the very presence of the predetermined gold standard, possibly because humans evaluators are anchored on that standard value of 100?

\textit{Anchoring bias}, which is the tendency of people to focus on the first piece of information presented; also defined as ``\textit{inability of the people to make sufficient adjustments starting from the initial value (anchor) to yield the final answer}'' \cite{kahneman2003perspective}. Decades of research has resulted in a robust finding that humans are prone to cognitive biases when engaged in decision-making \cite{tversky1974judgment,kahneman2003perspective,furnham2011literature,cho2017anchoring,wesslen2019investigating,dimara2018task}, which are heuristics that help humans reach decisions quickly \cite{kahneman2003perspective}. To the best of our knowledge, the impact of anchoring bias when humans evaluate conversational agent output has not heretofore been studied,  even as human evaluation has become an integral piece of most current research evaluating chatbots.

To investigate the effects of cognitive biases, specifically anchoring bias, on decision-making around evaluating chatbot output, we designed a 2$X$2 experiment with 77 crowdsourced workers. We studied how anchors (both numerical and textual) and the presentation order of rating tasks affect consistency of human judgments. We elicited ratings from workers on two metrics: Readability and Coherence of model output. Our key findings are listed below.
\begin{itemize}[nolistsep,noitemsep]
    \item We find systematic effects of anchoring in the \textbf{magnitude} of participants' ratings: participants who are presented with an anchor will provide a rating that is closer to the anchor value than those who are not presented with an anchor.
    \item We find systematic effects of anchoring in the \textbf{consistency} of participants' ratings: participants who are presented with an anchor will be (generally) more consistent in their ratings than those who are not presented with an anchor.
    \item We find that interpretation of metrics affects consistency: participants were more consistent with their ratings on Readability than in their ratings on Coherence, potentially because the interpretation of Coherence is more subjective than Readability.
\end{itemize}

Our findings demonstrate the impact anchoring bias might have in designing evaluation experiments. Along with exploring the impact of anchoring, we also provide insights into how the prior experience of being involved in similar research studies as well as time taken to complete the task as factors that can affect rating consistency. Our findings have the potential to advance the field of human-agent interaction by extending the reproducibility of conversational agent evaluation experiments. The findings of this paper are applicable to other areas of natural language processing, including text summarization and story generation, that also rely on human evaluation to study the quality of the algorithms. More broadly, the design of experiments in this paper can be adapted to investigate the effects of cognitive biases in a range of human-computer interaction tasks, building upon prior work in Explainable AI \cite{wang2019designing}  and bias mitigation \cite{sperrle2019human}.

\section{Related Work}
Our work relates to three primary areas of research; we present related work in each area in this subsection. 
\subsection{Cognitive Biases in Decision Making}
Evaluating algorithm output is an inherently subjective task. Cognitive biases, simple heuristics that are effective but may lead to suboptimal decision-making, especially when uncertainty is involved \cite{tversky1974judgment}, are a critical concern but surprisingly understudied when evaluating conversational agent output. Cognitive biases were first introduced by Tversky and Kahneman, and have been studied extensively in the field of psychology \cite{kahneman1972subjective,tversky1974judgment, ellis2018cognitive}. One form of cognitive bias is anchoring bias, which is when humans rely on a single piece of information (``anchor'') to make a decision \cite{kahneman201636}. Tversky and Kahneman \cite{tversky1974judgment} found evidence that when individuals are asked to provide an estimate, their estimates were pretty close to the reference value or anchor. Anchoring can thus affect decision-making in visual analytics \cite{cho2017anchoring,wesslen2019investigating}, valuations \cite{ariely2003coherent}, even general knowledge \cite{gilovich2001putting,furnham2011literature}. For Natural Language Processing tasks however, there has been little research studying the impact of anchoring bias. One prior study by Berzak \emph{et al.} \cite{berzak-etal-2016-anchoring} evaluated the impact of anchoring bias in the creation of syntactic parsers. When it comes to evaluating the output of conversational agents, there has been no prior work on understanding the impact of cognitive biases. Our work is the first step in that direction.

\subsection{Evaluation of Dialogue Systems} 
There are two main domains in which conversational agents are deployed: open-domain \cite{dziri2018augmenting,serban2016building,asghar2018affective} and goal-oriented  \cite{lipton2018bbq} conversational settings. Goal-oriented systems are designed to achieve a specific goal, such as restaurant booking \cite{bordes2016learning} or movie ticket booking \cite{li-etal-2017-end}.  Open-domain systems, also known commonly as chit-chat systems, engage with a conversation partner towards no predetermined goal \cite{zhang-etal-2018-personalizing}. Typically, natural language generation in conversational agents is achieved by training \textit{seq2seq} architectures \cite{vinyals2015neural,sutskever2014sequence}. 
Prior research has shown that agents built using \textit{seq2seq} frameworks suffer from generating dull and generic responses \cite{vinyals2015neural,li-etal-2016-diversity}.   
Evaluating the quality of responses generated by these models in open-domain situations is thus an important area of research because it affects user satisfaction and engagement \cite{walker1997paradise,venkatesh2018evaluating}.

To evaluate output \textit{automatically}, researchers have adopted metrics such as BLEU \cite{papineni2002bleu}, METEOR \cite{banerjee2005meteor} and ROUGE \cite{lin2004rouge} from machine translation and text summarization \cite{dialogue-eval} tasks. BLEU, METEOR and ROUGE can be computed based on word overlap between the proposed and ground truth responses; however, they do not adequately account for the diversity of responses that are possible for a given input utterance. Experiments show that these automated metrics along with word embedding based metrics \cite{dialogue-eval} show little to no correlation with human ratings \cite{dialogue-eval,lowe-etal-2017-towards}. With the lack of proper automated metrics for evaluation, obtaining human ratings is a primary evaluation method for evaluating chatbots. Even with human evaluation, a variety of metrics have been proposed, including \textit{ease of answering} \cite{Li-reinforce}, \textit{coherence} \cite{Li-reinforce}, \textit{information flow} \cite{Li-reinforce} , \textit{naturalness} \cite{asghar2018affective}, \textit{fluency} \cite{zhang-etal-2018-personalizing} and \textit{engagement} \cite{venkatesh2018evaluating}. Our current study builds upon this prior research and seeks to investigate the use of appropriate metrics in evaluating chatbots.  As an experiment design choice, we also asked crowdsourced workers which metrics they would themselves consider most important while undertaking these tasks. 

\subsection{Experiment Design in Language Evaluation}
Our focus in this paper is on experiment design. 
Our motivation to do so is based on prior research that demonstrated the effectiveness of different questions types (e.g. continuous scales, magnitude estimation, etc.) to obtain human ratings instead of using discrete scales (e.g. Likert scales) \cite{novikova-etal-2018-rankme,belz2011discrete,belz2010comparing,kiritchenko-mohammad-2017-best}. Likert Scales are widely used to obtain human ratings for conversational agent output \cite{dinan2018wizard,zhang-etal-2018-personalizing,chen2018hierarchical}. However, Likert scales suffer from a number of limitations such as inconsistencies in ratings by different annotators, scale region bias and fixed granularity \cite{kiritchenko-mohammad-2017-best,schuman1996questions,baumgartner2001response}. Recent work done by Novikova \emph{et al.} \cite{novikova-etal-2018-rankme} addresses the issue of inconsistency in ratings, although in goal-oriented systems. Their work demonstrates the effectiveness of using continuous scales towards increased consistency for language evaluation tasks. However, the extent to which anchoring bias may affect consistency has not been previously studied. Prior research from Novikova \emph{et al.} \cite{novikova-etal-2018-rankme} also demonstrates an increase in consistency when the rating tasks are split so that each metric is rated individually (rating Readability followed by rating Coherence). Taking inspiration from this, our experiment design has explicit conditions to investigate the effects of splitting the rating tasks.

To summarize this Related Work section, evaluation of dialogue system output relies increasingly on human evaluation, yet not a lot of research focuses on experiment design for this task. Also, we find very little work towards understanding the impact of cognitive biases that might affect ratings obtained from crowd-sourced workers. Our present study seeks to fill this research gap and propose better experiment design procedures for use by fellow researchers in this area.

\section{Corpus and Models}
To obtain ratings on conversational agent output, we trained three models from scratch to generate responses. Code for these models was made available by Dziri \emph{et al.} \cite{dziri2018augmenting} (\url{https://github.com/nouhadziri/THRED}). We first describe the corpus we used to train the models.
\subsection{Corpus}
We used the Reddit Conversational Corpus made available by Dziri \emph{et al.} \cite{dziri2018augmenting}. This corpus consists of conversations obtained from 95 different subreddits, curated out of 1.1M subreddits. The date range is a 20-month period from November, 2016 until August, 2018. Table~\ref{datasetstats} shows overall descriptive statistics of the corpus, where the average length of utterances is consistent across the Training, Validation and Test sets.

\begin{table}[htp]
\centering
\begin{tabular}{cccc}
\toprule
\textbf{}           & \textbf{Train} & \textbf{Valid.} & \textbf{Test} \\
\toprule
\textbf{Dialogues}  & 9.2M           & 500K            & 400K          \\ \midrule
\textbf{Avg. Length of Utterances} & 13.98         & 13.98           & 13.99         \\  \bottomrule
\end{tabular}
\caption{Descriptive statistics of the corpus used in our experiments.}
\label{datasetstats}
\end{table}

\subsection{Models}
\begin{figure*}[t]
  \centering
    \includegraphics[width=1\textwidth]{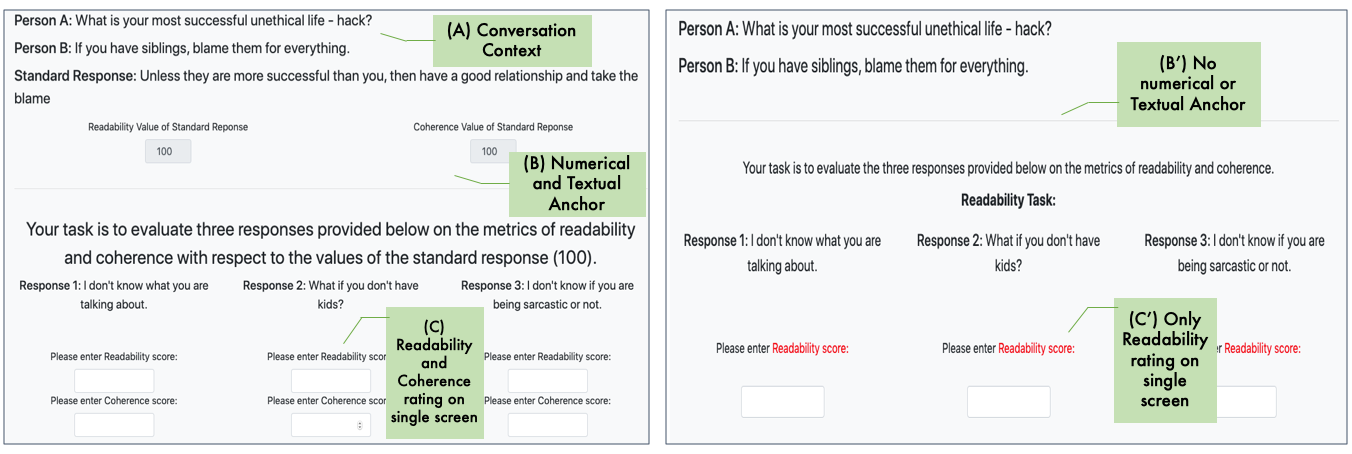}
  \caption{Sample screen showing variations in the experiment conditions. (A) represents the conversational context that is shown across all conditions. (B) is the numerical and textual anchor presented to participants in anchoring conditions. (B') shows the screenshot of conditions where no anchor is presented. (C) is used in Setup 1 where both questions of readability and coherence ratings are shown together. (C') is used in Setup 2 where the readability and coherence are treated as individual tasks and only one is shown at a time to the participant.}
  \label{fig:experiment_design_revised}
\end{figure*}

All three models used in our experiments are based on \textit{seq2seq} approaches that contain an encoder and decoder component. \textit{Seq2seq} approaches are commonly used in language generation tasks, such as machine translation and dialogue generation. For dialogue generation, the encoder receives the input sequence $X=x_1,x_2,....,x_n$ as input. Each input sequence is passed through an LSTM \cite{hochreiter1997long} on the encoder side which produces a hidden state representation (Eq \ref{encoder}.) 
\begin{equation}
    h_{t}^{enc} = f(h_{t-1}^{enc},x_t).
    \label{encoder}
\end{equation}
where $h_{t-1}^{enc}$ represents the previous hidden state and $f$ represents a non-linear activation function.
The decoder uses the last hidden state of the encoder as the initial state and output tokens are conditioned on the input (Eq \ref{decoder}.) where $y_{t-1}$ represents the ground truth input into the decoder.
\begin{equation}
    s_{t}^{dec} = f(s_{t-1}^{dec},y_{t-1})
    \label{decoder}
\end{equation}
\begin{enumerate}[nolistsep,noitemsep]
    \item \textbf{Seq2Seq:} Our first model is a traditional \textit{seq2seq} model with attention mechanism. We use the attention mechanism proposed by Bahdanau \emph{et al.} \cite{bahdanau2014neural}. Attention assists the decoder to attend to different parts of the input while generating the response. The decoder produces a context vector $c_t$ at each time step by attending to the encoder hidden state $h_{t}^{enc}$ along with the last hidden of the decoder $s_{t-1
    }$  (represented through Eq \ref{attention}.)     where $\alpha$ represents the relative importance on the input side. The output from the model $y_t$ is produced through a softmax function (Eq~\ref{softmax}.). 
    \begin{equation}
        \begin{split}
            c_i = \sum_{i=1}^{n} \alpha_{i}h_{i}^{enc} \\ 
            \alpha_{i} = \dfrac{exp(e_{i})}{\sum_{j=1}^{n} exp(e_{j})} \\
            e_{i} = f(s_{t-1},h_i)
        \end{split}
        \label{attention}
    \end{equation}
    \begin{equation}
        y_t = softmax(y_{t-1},s_t,c_t)
        \label{softmax}
    \end{equation}
    \item \textbf{HRED:} Our second model uses \textit{\textbf{Hierarchical Encoder-Decoder}} \cite{serban2016building} architecture. This model is an advancement over traditional \textit{seq2seq models}. HRED overcomes the bottlenecks of traditional \textit{seq2seq} models by capturing longer context from dialogue histories. HRED model introduces a two-level hierarchy to capture long term context. The first layer is called the utterance layer that captures the meaning of each sentence, similar to traditional seq2seq models. It further encodes the hidden states of the utterance layer to the inter-utterance layers that capture the context and input information \cite{tian-etal-2017-make}.
    \item \textbf{THRED:} Our last model is the \textit{\textbf{Topic Augmented Hierarchical Encoder-Decoder}} \cite{dziri2018augmenting}. This model uses topic words along with a hierarchical encoder-decoder to produce a response. The topic words were obtained using a pre-trained LDA model \cite{hoffman2010online}. This model also makes uses of attention mechanism on the context along with the topic words from the input sequence.  
\end{enumerate}
\textbf{Sample Output from Models:}
In Figure~\ref{fig:experiment_design_revised} (top-left), we show a sample conversation from the Reddit Corpus. It consists of two sentences, spoken by Person A and B. The corpus also provides the target (or gold-Standard) response against which the model can be trained, and also against which performance can be evaluated. This is shown in the Standard Response in Figure~\ref{fig:experiment_design_revised} screenshot. In the bottom of the screenshot, the output from the three generative models described in this section is shown (\textit{seq2seq}, HRED and THRED output in Response 1, 2 and 3 respectively). 

\section{Experiment Design}

Having obtained the outputs from our three models, we built an interface to allow participants to evaluate the generated responses. We initially focus on two metrics:  \textbf{Readability} and \textbf{Coherence}. Readability and Coherence are frequently used in obtaining evaluation ratings from crowd-sourced workers \cite{novikova-etal-2017-need,ghazvininejad2018knowledge,zhang-etal-2018-personalizing,asghar2018affective}. \textbf{Readability} measures the linguistic quality of text and helps quantify the difficulty of understanding the text by the reader \cite{Gatt:2018:SSA:3241691.3241693,novikova-etal-2017-need}. \textbf{Coherence} measures the ability to produce responses consistent with the topic or context of conversation \cite{venkatesh2018evaluating}. Based on prior findings of the limitations of Likert scales \cite{belz2011discrete},  
 
we instead use magnitude estimation (ME) questions to obtain ratings from crowdsourced workers. \textbf{Magnitude Estimation} allows participants to rate the responses over a free scale without being constrained. Recently, Novikova \emph{et al.} \cite{novikova-etal-2018-rankme} demonstrated that use of magnitude estimation helps improve consistency amongst crowd-sourced workers when evaluating responses from goal-oriented systems. We build upon this prior work but specifically focus on investigating the impact of cognitive biases to design our experiments. 

Accordingly, we design four experiment conditions, namely \textbf{Anchor}: With or Without Anchor and \textbf{Presentation Order}: Both Questions or Single Question (on a single screen). Table~\ref{distribution} shows the four different experiment conditions in our experiment design, while Figure~\ref{fig:experiment_design_revised} shows two sample screenshots from the study interface.

\begin{table}[h]
\centering
\begin{tabular}{ccc}
\toprule
                 & \textbf{No Anchor} & \textbf{Anchor} \\ \toprule
\textbf{Both Questions (Setup 1)} & 18                 & 22              \\ \midrule
\textbf{Single Question (Setup 2)} & 18                 & 19              \\ \bottomrule
\end{tabular}
\caption{$2X2$ experiment design with four experiment conditions and number of participants across each condition}
\label{distribution}
\end{table}

As shown in Figure~\ref{fig:experiment_design_revised}, participants across all experiment conditions are shown the Conversation Context (A). Participants in the Anchor conditions are shown the Standard Response and the Readability and Coherence value of the Standard Response (set to 100 in this study, following prior work done by \cite{novikova-etal-2018-rankme}); together these form the Numerical and Textual Anchor (B) (Figure~\ref{fig:experiment_design_revised}-left). Participants in the No Anchor condition are shown neither the Standard Response nor the Readability and Coherence value of the standard response (B') (Figure~\ref{fig:experiment_design_revised}-right). Participants in the Both Questions (Setup 1) condition are asked to input their ratings of Readability and Coherence on a single screen (C) (as shown in Figure~\ref{fig:experiment_design_revised}-left). Participants in the Single Question condition (Setup 2) are asked to input their ratings on a single metric on single screen (as shown Figure~\ref{fig:experiment_design_revised}-right (C') for Readability), and then input their ratings on the Coherence metric on the next screen when they click the next button (not shown). 

\begin{figure}[h]
  \centering
      \includegraphics[width=1\columnwidth]{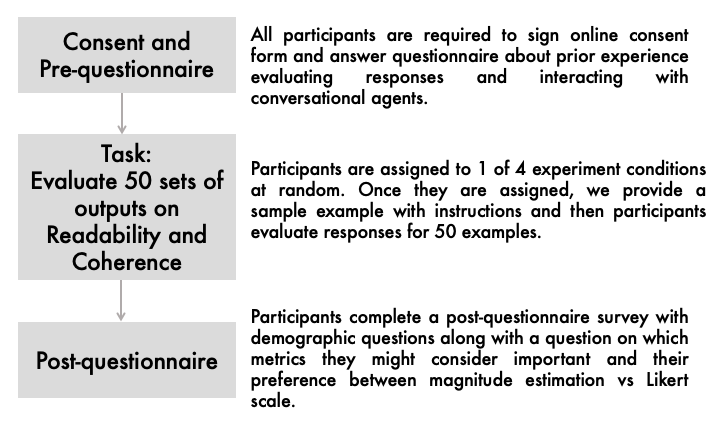}
  \caption{The experiment flow for each crowd-sourced worker taking part in this study.}
  \label{fig:overview}
\end{figure}

Figure \ref{fig:overview} provides the flow of steps taken by workers in the experiment, beginning with the informed consent procedure and pre-questionnaire, followed by the task of evaluating 50 sets of outputs on two metrics of Readability and Coherence and ending with the post-questionnaire. In the pre-questionnaire, we asked two questions about the prior experience of workers: (Q1) \textit{Have you taken part in previous studies that involve evaluating conversational responses?} and (Q2) \textit{Have you taken part in previous studies that involve talking to a chatbot?} Our motivation behind asking these questions is to understand if prior experience participating in similar studies affects inter-rater consistency. In the post-questionnaire, we obtain participant demographics including their age, gender, race, and education. We also ask them if they find it preferable to provide ratings as magnitude estimation question or on Likert scales. In addition, we obtain their free-form responses on which metrics they would consider important for evaluating conversational agent output. These post-questionnaire questions are designed to obtain qualitative data to better inform our future studies.

\subsection{Research Questions}
Following the review of prior work in this area and our decisions on the experiment design, we developed three main research questions for our study. 
\begin{itemize}[nolistsep,noitemsep]
    \item \textbf{RQ1:} Which factors affect the \textit{magnitude} of ratings provided by the participants?
    \textbf{Rationale:} The presence of an anchor may orient participants towards that number (100) and also the reference text, thus we expect that participants in the anchoring conditions will have higher ratings (closer to 100) than do participants in the no anchor conditions. In addition, we investigate if the presentation order of questions (Setup 1 vs. Setup 2) has an effect on how high participants' ratings are on the task. We also investigate whether the time to complete the task has any effect on the magnitude of ratings. We use the responses on the pre-questionnaire about the prior experience to analyze whether having taken part in similar studies or conversing with a chatbot has any effect on the magnitude of ratings.  

    \item \textbf{RQ2:} Which factors affect the \textit{consistency} in ratings provided by participants?   \textbf{Rationale:} Similar to RQ1, we expect that the presence of an anchor may orient participants towards that number (100) and also the reference text, thus we expect that participants in the anchoring conditions will have higher consistency in their ratings than do participants in the no anchor conditions. In addition, we investigate if the presentation order of questions (Setup 1 vs. Setup 2) has an effect on the consistency of ratings on the task. We also investigate whether the time to complete the task and prior experience affect inter-rater consistency. 
    \item \textbf{RQ3:} Are participants more consistent in their ratings of readability than coherence? \textbf{Rationale:} Across both setups, we except higher consistency in readability ratings than coherence. We also expect the impact of anchoring to be more pronounced for readability over coherence. We contend coherence is more subjective to evaluate than is readability, since humans have judge whether the response is related to the context of the conversation \cite{dziri-etal-2019-evaluating-coherence,novikova-etal-2017-need}. Readability on the other hand has been evaluated across other fields through automated metrics and is more well-defined  \cite{kincaid1975derivation}.  
\end{itemize}

\section{Results}
We present the results of our analysis in this section. We begin by describing the pool of participants we recruited and the quality checks we put in place to ensure high-quality crowdsourced data. 

\subsection{Descriptive Statistics}
Our study was approved under our institution's Institutional Review Board (IRB) policies (IRB \#18-0357). Table \ref{distribution} provides the number of participants across the four experiment conditions. We recruited crowdsourced workers through Amazon Mechanical Turk.\footnote{\url{https://www.mturk.com/}} The participants were assigned to experiment condition randomly.  We allowed each participant a maximum of 4 hours to complete the study. In order to ensure high-quality data, we had stringent qualifying criteria: (1) Workers should have a Masters qualification;\footnote{\url{https://www.mturk.com/worker/help}} (2) HIT Approval Rate to be $>$ 80; and (3) Number of approved HITs $>$ 500. Our study interface was hosted on a secure server at our institution and participant responses were saved in a MongoDB database. In our analyses, we use the aggregated value of the responses provided by each participant for the entire task. The entire anonymized data, analyses and code used in this study are available at this link: \url{https://github.com/sashank06/ConvEvaluation_CHI2020}.

A total of 77 crowdsourced workers participated in our study. The gender distribution was 67.5\% male (52), 31.17\% female (24) and 1.33\% other (1). The age of workers was between 20 and 60 years (mean=34.85 years). A majority of the participants had an undergraduate degree ($n=37$), while others indicated having a Masters degree ($n=23$), Doctorate ($n=6$) and High School diploma ($n=11$). In terms of race/ethnicity, $n=37$ were Indian, along with White ($n=27$), Black ($n=2$), East Asian ($n=7$), Hispanic ($n=3$) and Native American ($n=1$) making up rest of the demographics.

In the pre-questionnaire, we also asked participants to indicate: (Q1) whether the participant has taken part in prior research studies evaluating conversational responses; and (Q2) interacting with a chatbot. Table~\ref{prior_experience} provides the number of participants' response across both setups to the pre-questionnaire questions.

\begin{table}[h]
\centering
\begin{tabular}{cccccl}
\toprule
                         &                      & \multicolumn{2}{l}{Question 1} & \multicolumn{2}{c}{Question 2} \\ \toprule
\multicolumn{1}{l}{}     & \multicolumn{1}{l}{} & Yes            & No            & Yes  & \multicolumn{1}{c}{No}  \\ \toprule
Setup 1                  & No Anchor            & 5              & 13            & 5    & 13                      \\ \cmidrule(l){2-6}
                         & Anchor               & 4              & 18            & 5    & 17                      \\ \midrule
\multirow{2}{*}{Setup 2} & No Anchor            & 7              & 11            & 8    & 10                      \\ \cmidrule(l){2-6}
                         & Anchor               & 6              & 13            & 7    & 12    \\ \bottomrule                 
\end{tabular}
\caption{Number of participants in each category: we refer to prior experience on evaluating conversational output as Question 1 and prior experience of engaging with chatbots as Question 2.}
\label{prior_experience}
\end{table}

\subsection{Analysis and Results for RQ1}

\begin{figure}[h]
  \centering
    \includegraphics[width=1\columnwidth]{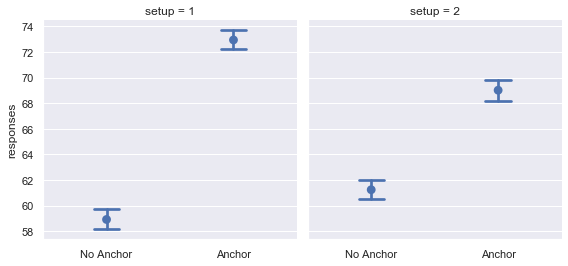}
  \caption{Mean of the responses bootstrapped with 95\% confidence intervals across setups 1 and setup 2} 
  \label{fig:primary}
\end{figure}

\begin{figure}[h]
  \centering
    \includegraphics[width=1\columnwidth]{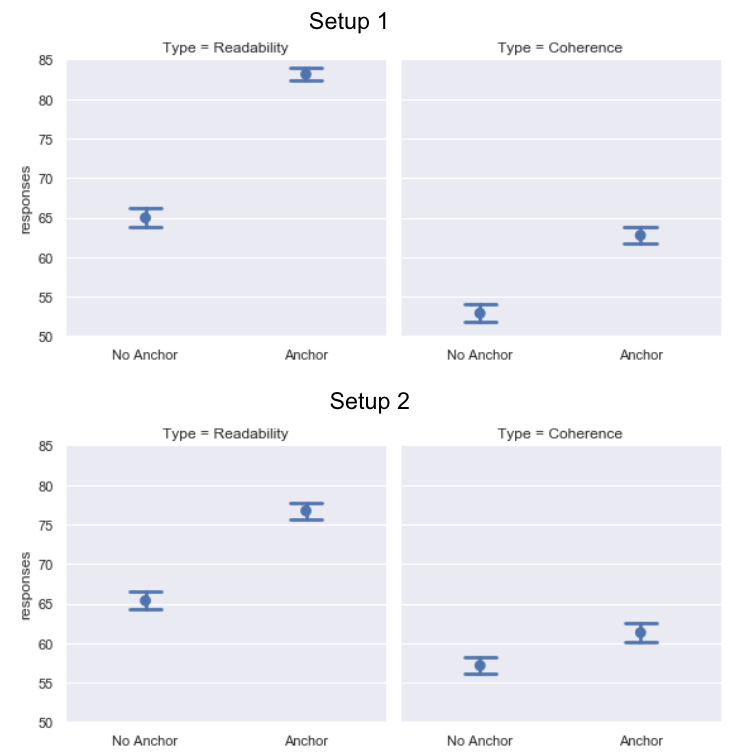}
  \caption{Mean of the responses bootstrapped with 95\% confidence intervals across Setups 1 and 2 on the metrics of Readability and Coherence.} 
  \label{fig:primary_finegrained}
\end{figure}

\paragraph{Effects of anchor and type of setup on magnitude of ratings}
We find significant differences between the magnitude of responses provided by participants across the both setups with p $<$ 0.001. Figure \ref{fig:primary} provides the mean and bootstrapped confidence interval (95\%) of the responses across the experiment conditions. In Setup 1, we find that participants with no anchor produce ratings ($M=58.92$) that are significantly lower than ratings provided by participants in anchor condition ($M=72.94$). We find a similar pattern across Setup 2 with no anchor, resulting in a mean rating of $61.25$, while ratings in anchor condition responses have a mean of $69.02$. We analyze the ratings on Readability and Coherence separately (Figure~\ref{fig:primary_finegrained}): the presence of numerical and textual anchors results in higher (on average) ratings than the absence of the anchor (statistically significant with p$<$0.001).  

Figure \ref{fig:primary_finegrained} presents ratings for the metrics of readability and coherence separately. We find that across both setups, the difference between anchor and no anchor conditions to be larger for the metrics of readability than coherence (statistically significant with p$<$0.001). We find that in Setup 1, readability values have a mean of 83.13 in the anchor condition and in no anchor condition the mean of the responses drop down to 64.97. Also in Setup 1, we find that for coherence metric, the mean of responses in the anchoring condition is M=62.74 and without anchor M=52.89. We find similar trends in the responses provided in Setup 2 for both metrics of readability and coherence.

\begin{figure}[h]
  \centering
    \includegraphics[width=1\columnwidth]{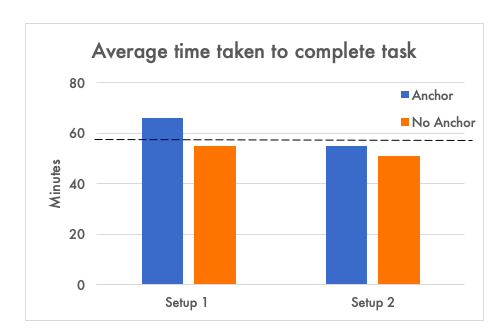}
  \caption{Average time taken to complete the task across four experiment conditions. Overall average is shown in dashed line in the graph (57.17 minutes).}
  \label{fig:time_taken}
\end{figure}

\begin{figure}[h]
  \centering
    \includegraphics[width=1\columnwidth]{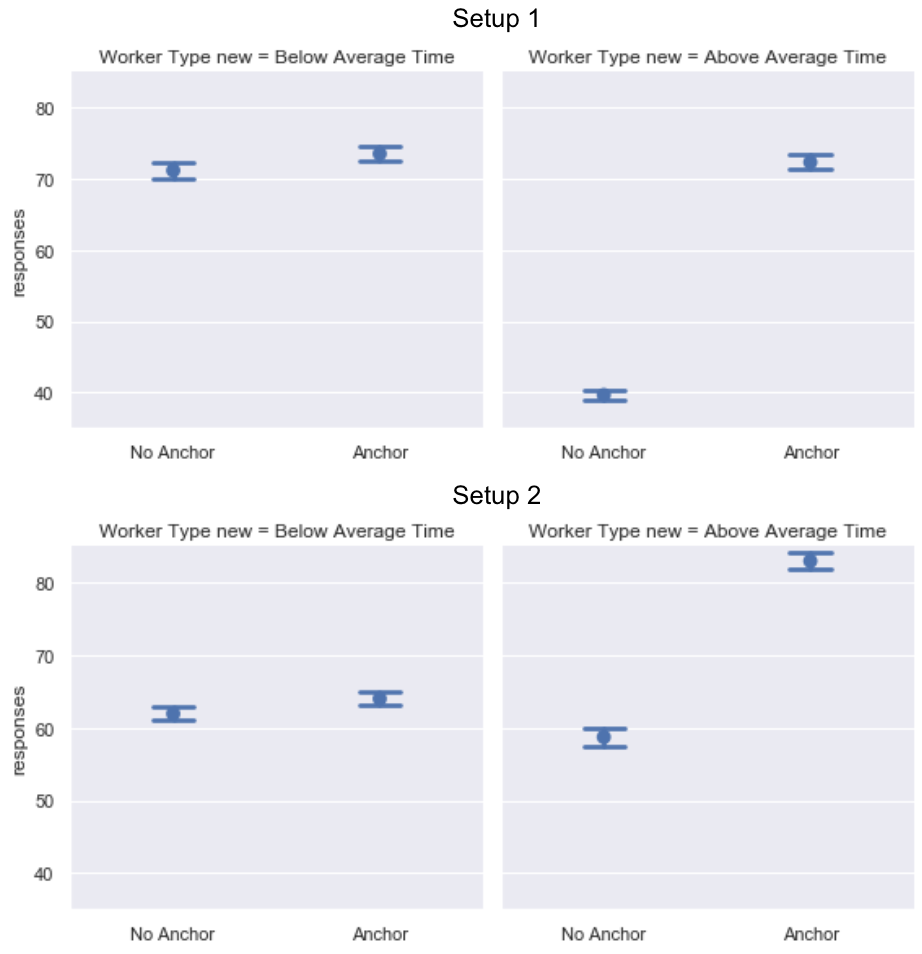}
  \caption{Mean of the responses bootstrapped with 95\% confidence intervals across Setups 1 and 2 based on amount of time spent on study.}
  \label{fig:external_time}
\end{figure}

\paragraph{Effect of time taken to complete task on magnitude of ratings}
We analyze the effects of time taken to complete the task on magnitude of ratings. We find that participants who are presented with anchors spend \textit{more time} on average taking the study than participants in no anchor conditions across both setups. From the total of 77 participants, the mean time taken to complete the study was 57.17 minutes (see Figure~\ref{fig:time_taken}). In Setup 1, we find that participants took an average of 66 minutes in the with anchor condition and average of 54.83 minutes in the without anchor conditions. Similarly, in Setup 2 we find participants took an average of 54.94 minutes with anchor condition and 50.94 with no anchor condition.

Next, we grouped the participants based on the amount of time spent into two categories: (1) \textbf{Below Average} - when participants spend less than mean time; (2) \textbf{Above Average} - when participants spend more than mean time. Table \ref{time_table} provides the number of participants based on the time spent across the experiment conditions. Across both setups, we find that people in the above average group show significant differences in their responses. In Setup 1, in the above average group, the mean of responses in no anchor condition was $39.65$ and mean of the responses in anchor condition was $72.35$. We find similar evidence in Setup 2 with people in anchor condition provide higher values ($83$) close to the numerical anchor ($100$). Although, we note that the sample sizes in the Below Average time taken groups in Setup 2 are smaller (4 and 5 participants resp., c.f. Table~\ref{time_table}); more experimentation is needed to further substantiate this finding.

\begin{table}[]
\begin{tabular}{@{}cccc@{}}
\toprule
 &  & Below Average & Above Average \\ \midrule
\multirow{2}{*}{Setup 1} & No Achor & 7 (71.19) & 11 (39.65) \\ \cmidrule(l){2-4} 
 & Anchor & 11 (73.53) & 11 (72.35) \\ \midrule
\multirow{2}{*}{Setup 2} & No Anchor & 4 (61.96) & 14 (58.75) \\ \cmidrule(l){2-4} 
 & Anchor & 5 (64.02) & 14 (83) \\ \bottomrule
\end{tabular}
\caption{Number of the participants who spent below and above average time across conditions and their average rating values (in parenthesis)} \label{time_table}
\end{table}

\begin{figure}[t]
  \centering
    \includegraphics[width=1\columnwidth]{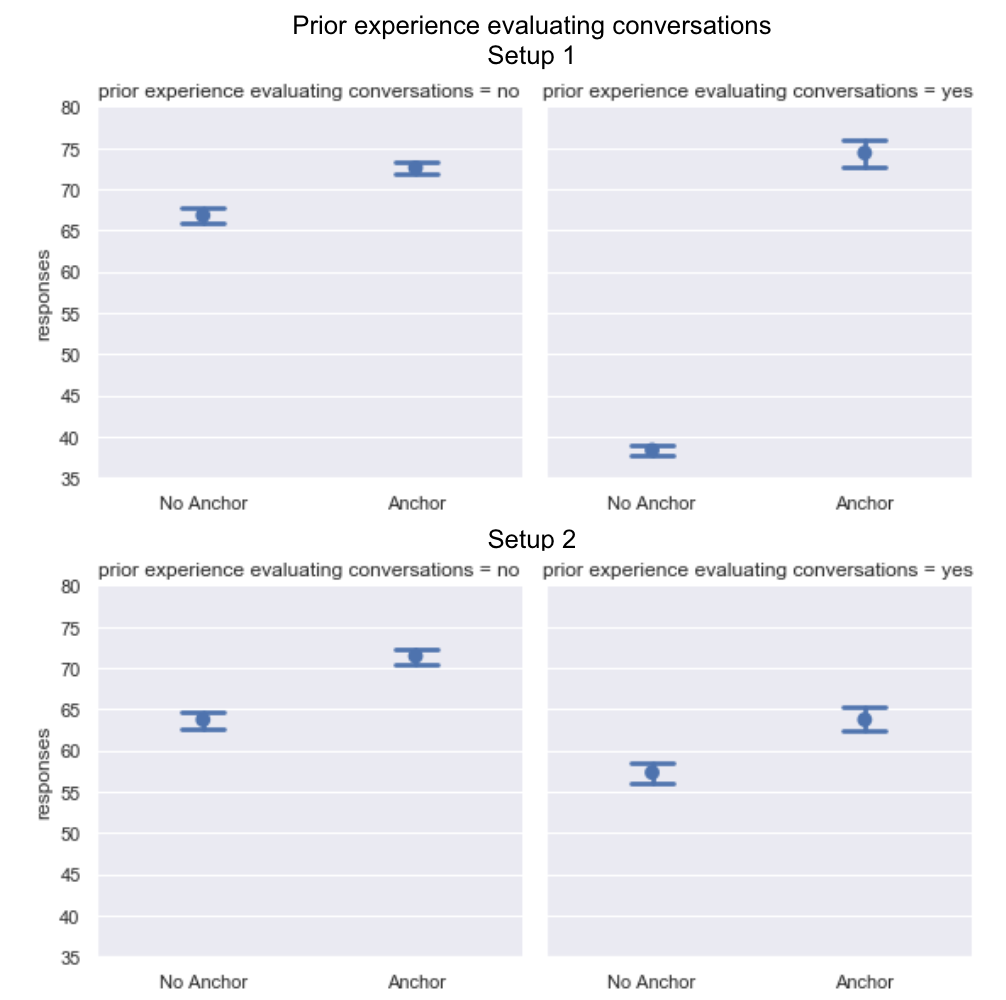}
  \caption{Mean of the responses bootstrapped with 95\% confidence intervals across setups 1 and setup 2 based on prior experience of being involved studies about evaluating conversations.}
  \label{fig:external_questions1}
\end{figure}

\begin{figure}[t]
  \centering
    \includegraphics[width=1\columnwidth]{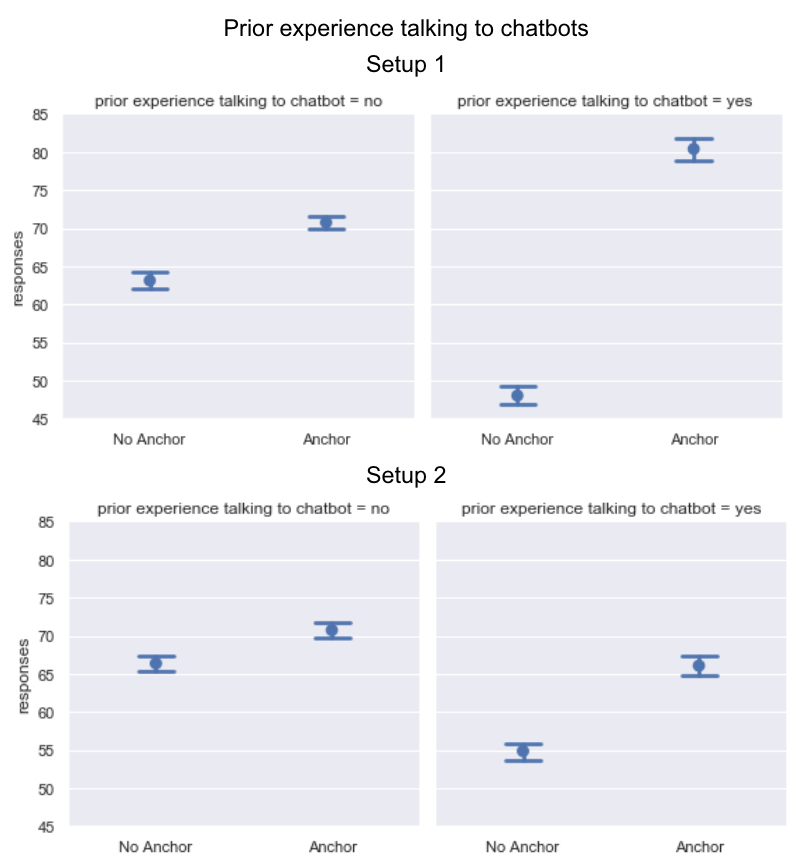}
  \caption{Mean of the responses bootstrapped with 95\% confidence intervals across setups 1 and setup 2 based on prior experience of being involved studies about talking to chatbot.}
  \label{fig:external_questions2}
\end{figure}

\paragraph{Effect of prior experience on magnitude of ratings}
Figure \ref{fig:external_questions1} demonstrates the impact of the prior experience of evaluating conversational responses (Question 1 on the pre-questionnaire) on the magnitude of ratings. We find contrasting responses across both setups. In Setup 1, we find that people with prior experience in the anchor condition produce higher responses (M=74.41) close to the numerical anchor (100) and no anchor condition produce lower values (M=38.36) whilst people with no prior experience are similar in their responses across both conditions. In comparison to Setup 1, we find that in Setup 2 participants with no prior experience produce higher responses in the anchor condition (M=71.45) and in no anchor condition (M=63.74).

Figure \ref{fig:external_questions2} shows the impact of prior experience of interacting with chatbots.  Participants who have such prior experience demonstrated signs of anchoring. We find that mean of responses (M=80.40) for participants with prior experience in the anchor condition to be significantly higher ($p<0.001$) than participants in no anchor condition (M=48.01) in Setup 1. 

When comparing against Setup 1, we find that people in Setup 2 with no prior experience produce higher responses (M=70.74) in the anchoring condition than in the no anchor condition (M=63.12). 

These findings substantiate the hypothesis that people with prior experience (answered Yes on Questions 1 and 2) would be more susceptible to the anchoring effect than those who do not have prior experience with similar tasks, \textbf{however} this effect is only seen in Setup 1, while Setup 2 demonstrates the opposite effect. We find this evidence to be particularly interesting and plan to further investigate the potential of eliciting ratings on different metrics as separate tasks (Setup 2) as a means of mitigating the anchoring bias effect.

\subsection{Analysis and Results for RQ2}
We  measure consistency of ratings using the intra-class correlation measure (ICC) \cite{landis1977measurement}. Following Bard \emph{et al.} \cite{bard1996magnitude}, we perform a log normalization of the scores obtained using magnitude estimation method across both setups.

\paragraph{Effects of anchor and type of setup on consistency of ratings}
Table \ref{icc-alldata} represents the ICC scores obtained across both setups on the metrics of readability and coherence. We find that there is a significant $(p<0.001)$ increase in the consistency of the ratings in the anchor condition in Setup 1. The consistency values obtained in Setup 2 for readability and coherence show mixed results. We find that the no anchor condition of Setup 2 produces more consistency in ratings for the readability metrics whilst on the metric of coherence, we find that there is extremely low consistency between the raters when they are presented with no anchors. However, we see a significant increase in consistency for Setup 2, when participants are in anchoring condition.

\begin{table}[h]
\centering
\begin{tabular}{cccc}
\toprule
                         &           & Readability & Coherence \\ \toprule
\multirow{2}{*}{Setup 1} & No Anchor (n=18) & 0.74        & 0.76      \\ \cmidrule{2-4}
                         & Anchor (n=22)    & 0.921       & 0.855     \\ \midrule
\multirow{2}{*}{Setup 2} & No Anchor (n=18) & 0.874       & 0.151     \\ \cmidrule{2-4}
                         & Anchor (n=19)    & 0.835       & 0.727   \\ \bottomrule
\end{tabular}
\caption{ICC scores on the metrics of readability and coherence for each experiment condition.  All values are statistically significant p-value$<$0.001}
\label{icc-alldata}
\end{table}

\begin{table}[h]
\begin{tabular}{@{}llll@{}}
\toprule
\textbf{Condition} & \textbf{Time Taken} & \textbf{Readability} & \textbf{Coherence} \\ \midrule
\multirow{2}{*}{\begin{tabular}[c]{@{}l@{}}Setup 1\\ No Anchor\end{tabular}} & \begin{tabular}[c]{@{}l@{}}Below Average\\ (n=11)\end{tabular} & 0.75 & 0.63 \\ \cmidrule(l){2-4} 
 & \begin{tabular}[c]{@{}l@{}}Above Average\\ (n=7)\end{tabular} & 0.23 & 0.59 \\ \midrule
\multirow{2}{*}{\begin{tabular}[c]{@{}l@{}}Setup 1\\ Anchor\end{tabular}} & \begin{tabular}[c]{@{}l@{}}Below Average\\ (n=11)\end{tabular} & 0.86 & 0.785 \\ \cmidrule(l){2-4} 
 & \begin{tabular}[c]{@{}l@{}}Above Average \\ (n=11)\end{tabular} & 0.83 & 0.68 \\ \midrule
\multirow{2}{*}{\begin{tabular}[c]{@{}l@{}}Setup 2\\ No Anchor\end{tabular}} & \begin{tabular}[c]{@{}l@{}}Below Average\\ (n=14)\end{tabular} & 0.85 & -0.03$\dagger$. \\ \cmidrule(l){2-4} 
 & \begin{tabular}[c]{@{}l@{}}Above Average \\ (n=4)\end{tabular} & 0$\dagger$. & 0$\dagger$. \\ \midrule
\multirow{2}{*}{\begin{tabular}[c]{@{}l@{}}Setup 2\\ Anchor\end{tabular}} & \begin{tabular}[c]{@{}l@{}}Below Average\\ (n=14)\end{tabular} & 0.726 & 0.76 \\ \cmidrule(l){2-4} 
 & \begin{tabular}[c]{@{}l@{}}Above Average\\ (n=5)\end{tabular} & 0.556 & -0.20$\dagger$. \\ \bottomrule
\end{tabular}
\caption{ICC scores on the metrics of readability and coherence based on the amount of time spent in the study across both conditions. All values statistically significant at p-value$<$0.001 except those indicated by $\dagger$.}
\label{icc_time}
\end{table}

\begin{table}[h]
\begin{tabular}{@{}llll@{}}
\toprule
\textbf{Condition} & \textbf{\begin{tabular}[c]{@{}l@{}}Prior experience\\ evaluating \\ conversations?\end{tabular}} & \textbf{Readability} & \textbf{Coherence} \\ \midrule
\multirow{2}{*}{\begin{tabular}[c]{@{}l@{}}Setup 1\\ No Anchor\end{tabular}} & Yes (n=5) & 0.44 & 0.71 \\ \cmidrule(l){2-4} 
 & No (n=13) & 0.67 & 0.62 \\ \midrule
\multirow{2}{*}{\begin{tabular}[c]{@{}l@{}}Setup 1\\ Anchor\end{tabular}} & Yes ((n=4) & 0.61 & 0.52 \\ \cmidrule(l){2-4} 
 & No (n=18) & 0.91 & 0.84 \\ \midrule
\multirow{2}{*}{\begin{tabular}[c]{@{}l@{}}Setup 2\\ No Anchor\end{tabular}} & Yes (n=7) & 0.77 & -0.88$\dagger$ \\ \cmidrule(l){2-4} 
 & No (n=11) & 0.71 & 0.46 \\ \midrule
\multirow{2}{*}{\begin{tabular}[c]{@{}l@{}}Setup 2\\ Anchor\end{tabular}} & Yes (n=6) & -0.2$\dagger$ & 0.65 \\ \cmidrule(l){2-4} 
 & No (n=13) & 0.93 & 0.86 \\ \bottomrule
\end{tabular}
\caption{ICC scores on the metrics of readability and coherence when based of participants prior experience of taking part in research studies about evaluating conversations. All values statistically significant at p-value$<$0.001 except those indicated by $\dagger$.}
\label{icc-q1}
\end{table}

\paragraph{Effect of time taken to complete task on consistency of ratings}
We look at the role of external factors of time and prior experience towards consistency of the ratings provided. Table \ref{icc_time} represents the ICC scores on the metrics on readability and coherence across both setups. We group these participants into two groups of \textbf{Above Average} and \textbf{Below Average}  based on the amount of time spent in the study (c.f Table \ref{time_table}). 

Surprisingly, we find that people who spend below average time achieve higher consistency in the ratings across both setups. However, we do notice some differences between the two setups. In Setup 1, we find that amongst participants who are in the below average group, the participants in the anchor condition have a higher consistency than participants in no anchor condition. Similarly, we find that people who spend above average time on Setup 1 with anchor condition achieve higher consistency when compared to Setup 1 with no anchor condition for the above average group. However, in Setup 2 we find people who spend above average time have a poor consistency score on the metric of coherence, a possible indication that coherence is highly subjective. 

\paragraph{Effect of prior experience on consistency of ratings}
Table \ref{icc-q1} provides an overview of the consistency on the readability and coherence metrics based on participants prior experience about taking part in studies about evaluating conversations across both setups. We find that participants with no prior experience of evaluating conversation across both setups tend to have higher consistency when compared to participants with prior experience of evaluating conversations irrespective on experimental condition assigned. When compared within the anchor conditions across both setups, we find that participants with no prior experience of evaluating conversations achieve higher consistency in Setup 2 and participants with prior experiences of evaluating conversations achieve a higher consistency on readability metrics with Setup 1. 

\begin{table}[]
\begin{tabular}{@{}llll@{}}
\toprule
\textbf{Condition} & \textbf{\begin{tabular}[c]{@{}l@{}}Prior experience\\ interacting with \\ chatbots?\end{tabular}} & \textbf{Readability} & \textbf{Coherence} \\ \midrule
\multirow{2}{*}{\begin{tabular}[c]{@{}l@{}}Setup 1\\ No Anchor\end{tabular}} & Yes (n=5) & 0.73 & 0.75 \\ \cmidrule(l){2-4} 
 & No (n=13) & 0.55 & 0.58 \\ \midrule
\multirow{2}{*}{\begin{tabular}[c]{@{}l@{}}Setup 1\\ Anchor\end{tabular}} & Yes ((n=5) & 0.89 & 0.69 \\ \cmidrule(l){2-4} 
 & No (n=17) & 0.87 & 0.79 \\ \midrule
\multirow{2}{*}{\begin{tabular}[c]{@{}l@{}}Setup 2\\ No Anchor\end{tabular}} & Yes (n=8) & 0.85 & -0.163$\dagger$ \\ \cmidrule(l){2-4} 
 & No (n=10) & 0.58 & -0.48 \\ \midrule
\multirow{2}{*}{\begin{tabular}[c]{@{}l@{}}Setup 2\\ Anchor\end{tabular}} & Yes (n=7) & -0.2$\dagger$ & 0.49 \\ \cmidrule(l){2-4} 
 & No (n=12) & 0.91 & 0.82 \\ \bottomrule
\end{tabular}
\caption{ICC scores on the metrics of readability and coherence when based of participants prior experience of taking part in research studies talking to chatbot. All values statistically significant at p-value$<$0.001 except those indicated by $\dagger$.}
\label{icc-q2}
\end{table}

Table \ref{icc-q2} gives an overview of the consistency on the readability and coherence metrics based on participants prior experience of taking part in studies related to engagement with a chatbot. Compared to Table \ref{icc-q1}, we find that participants with prior experience of engaging with chatbots achieve higher consistency across both setups irrespective of the experiment condition except on the Setup 2 anchoring condition. Also, we find the anchoring condition enables participants to achieve higher consistency across both Setup 1 and Setup 2. We find that irrespective of the participants' prior experience, anchoring helps achieve a higher consistency. This also provides similar evidence to presence of anchoring helping towards achieving higher consistency in this experiment design. Tables with confidence intervals for Figures 3, 4, 6, 7 and 8 are included in our github repository.

\subsection{Analysis and Results for RQ3}
As shown in Table~\ref{icc-alldata}, we see that readability has a higher consistency over coherence on both setups. We also notice the significant impact anchoring has towards increasing consistency of ratings. We see that it seems harder to agree upon the more  subjective metric of coherence, without any textual or numerical anchor. We also suspect the impact of instructions might have towards consistency. In the instructions screen in our study, Readability was defined as: \textit{Is the response easy to understand, fluent and grammatical and does not have any consecutive repeating words} (following \cite{novikova-etal-2017-need, novikova-etal-2018-rankme}), which provides clear indicators regarding evaluating a response on the metric of readability. Coherence was defined as: \textit{Is the response relevant to the topic and context of the conversation.} (following \cite{dziri2018augmenting,venkatesh2018evaluating} making it more subjective.

\section{Discussion and Limitations}
In this section, we discuss implications of our results on anchoring effect in dialogue evaluation, and point out possible limitations related to the study design and analysis.

\subsection{Implication of experiment results}
Our key findings indicate that the presence of numerical and textual anchors significantly influences the ratings across two different experiment setups. We find the effect of anchoring is more pronounced in instances when participants are asked to provide ratings on two metrics at the same time (Both Questions/Setup 1) and the effect of anchoring is slightly less pronounced when participants are asked to provide ratings for a single metric on a single screen (Single Question/Setup 2). Our findings have implications for potential future experiment designs that are geared towards evaluating the performance of dialogue systems, if there are ratings to be elicited on multiple dimensions, such as Readability and Coherence. 

Additionally, external factors of time taken to complete the study and participants prior experience of having taken part in research studies either about evaluation or engagement with a chatbot were found to impact the magnitude of the responses and consistency in the ratings. We find participants who spend more than the average time (above average) on the study get anchored and also exhibit low consistency scores on the metrics of readability and coherence. 

We notice the choice of metrics to evaluate also has an impact on consistency. We see that ratings for the more subjective metric of coherence are less consistent than those for readability amongst the raters across all conditions and setups.

We also analyzed the data from the post-questionnaire questions asking participants which method of rating they preferred to work with. From the 77 participants, we find that 42 participants preferred the magnitude estimation method and 35 of them preferred the Likert scale method. Prior research has shown that continuous scale methods like magnitude estimation do offer advantages  \cite{belz2011discrete,novikova-etal-2018-rankme} and they need to be explored further for the purposes of evaluation. Consistent with the prior work in this area, we also find similar advantages provided by magnitude estimation across both our setups with an increase in consistency of the ratings provided by the crowd-sourced workers. These findings and the participants' feedback on their own preferences lead us to recommend magnitude estimation for future evaluation design of conversational agents. 

\paragraph{Limitations}
We acknowledge a few limitations of our work. First, we consider only two metrics for evaluation of conversational agents. In reality, there may be more metrics that are better designed to evaluate the performance of conversational agents. Second, we acknowledge that this study is exploratory; understanding the impact of anchoring bias in the evaluation of conversational agents is in its infancy. For future studies, we plan to pre-register our study to improve the validity of our findings \cite{kosara2018skipping}. Third, we study the effect of anchoring, however, we provide both numerical and textual anchors. Although we find the impact of anchoring, we are unable to determine if the numerical or the textual anchor is causing this effect. To address this, we are planning an extension study with additional experiment conditions so that we can study the impact of textual and numerical anchors separately. 

\paragraph{Future Work}
\begin{figure}[t]
  \centering
    \includegraphics[width=1\columnwidth]{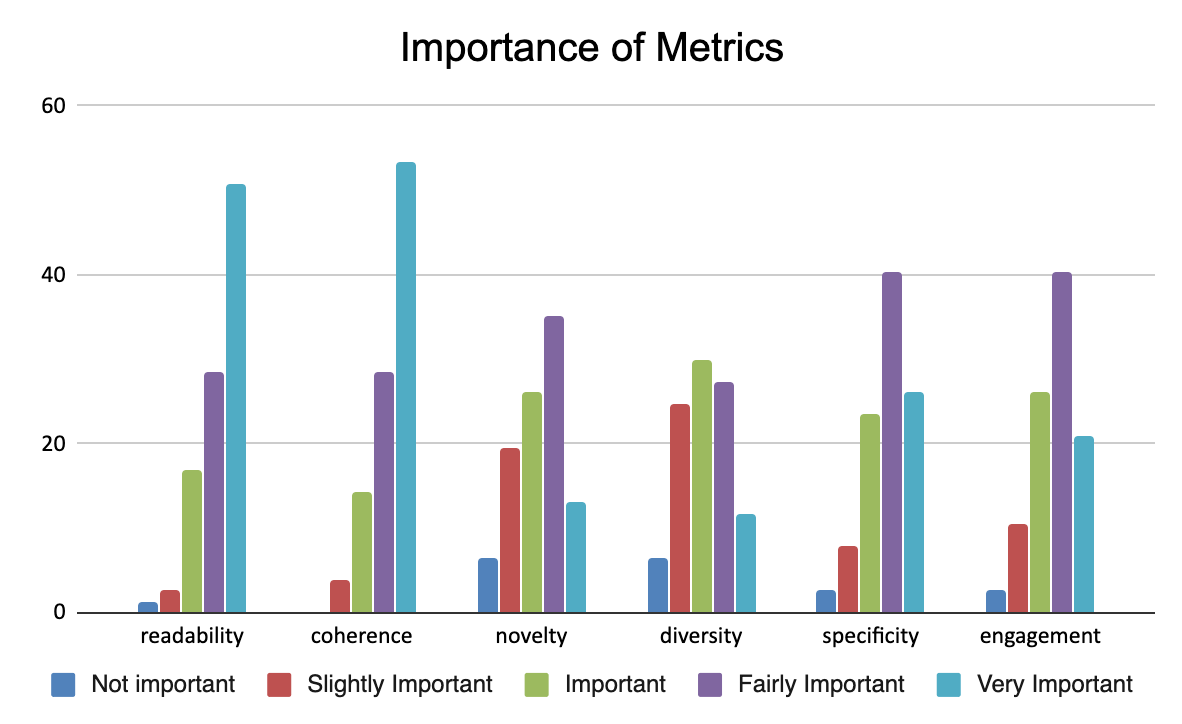}
  \caption{Participant ratings on which metrics they considered important for conversational output evaluation. Y-axis represents the \% of importance.}
  \label{fig:fw}
\end{figure}
The results of our study offer insights into the challenging task of designing and understanding the impact of experiments for evaluation of dialogue systems. To provide additional information for possible future directions, we also asked the participants in our study to rank the metrics that they considered important for output of conversational agents (Figure \ref{fig:fw}), including Readability and Coherence. We ask them to rate their preferences in order of importance for the following metrics: Readability, Coherence, Novelty, Diversity, Specificity and Engagement. These metrics are some of the commonly used metrics in research articles that develop and evaluate conversational agent output. We notice that readability and coherence are considered very important, but other metrics such as engagement and specificity are also worth investigating. Possible extensions to our work would include specificity and engagement metrics, based on this evidence. Past research by See \emph{et al.} \cite{see-etal-2019-makes} specifies metrics including specificity and engagement/interestingness  and shows how these metrics could impact the training process of a model.

\section{Summary}

Evaluation of dialogue systems is an extremely challenging task since automated metrics do not adequately capture the nuances related to natural language and its production. However, prior research has not focused on the impact that experiment design has on qualitative dialogue evaluation. 

Our findings are a step towards understanding the impact of experiment design and the possible role of cognitive bias such as anchoring bias towards dialogue evaluation. Cognitive biases could be the result of System 1 thinking (Type 1 processing), which is considered to be relatively fast, relatively low on cognitive demand, often based on intuition. By contrast, System 2 thinking (or Type 2 processing), is considered to be the result of systematic thinking and reasoning. Our results, however, indicate that participants who spent less time on the task had higher consistency of ratings than those who took longer. One possible experiment to identify the effects of Type 1 vs. Type 2 processing is to design an experiment condition which explicitly triggers intuitive responses (Type 1) by imposing a strict and challenging response deadline. Bago and De Neys \cite{bago2017fast} observed in their experiments that participants gave correct, logical responses as the first, immediate response, by explicitly triggering Type 1 vs. Type 2 processing for logic problems. Capturing time taken per question in the interaction logs would allow us to collect the data that supports this investigation.

We specifically investigate impact of anchoring bias in our experiment, to determine its effects on the consistency measure across participants. By separately analyzing the effect of the presence/absence of anchors and also the presentation order of questions, we are able to make design recommendations for future experiments on dialogue evaluation. We focus on the metrics of readability and coherence, but our proposed experiment design can be extended to multiple other metrics. In addition, our study also suggests that external factors of time and prior experience of taking part in research studies about evaluation of responses and engagement with chatbots have a significant impact towards responses provided and also on consistency.

 \section*{Acknowledgments}
This work was supported by the Defense Advanced Research Projects Agency (DARPA) under Contract No FA8650-18-C-7881. All statements of fact, opinion or conclusions contained herein are those of the authors and should not be construed as representing the official views or policies of AFRL, DARPA, or the U.S. Government. We thank the anonymous reviewers for the helpful feedback.

\bibliographystyle{SIGCHI-Reference-Format}
\balance
\bibliography{sample}

\end{document}